\documentclass[manuscript]{acmart}

\usepackage[utf8]{inputenc}
\usepackage{blindtext}
\usepackage{caption}
\usepackage{array}
\usepackage{booktabs}
\usepackage{tabularx}
\usepackage{xcolor}
\usepackage{enumitem}
\usepackage{balance}
\usepackage{times}
\usepackage{graphicx}
\usepackage{amsmath}
\usepackage{multicol}
\usepackage{multirow}
\usepackage{booktabs}
\usepackage{enumitem}
\usepackage{overpic}
\usepackage{pifont}
\usepackage{balance}
\usepackage{arydshln} 
\usepackage{MnSymbol,wasysym}
\usepackage{sidecap, caption}

\definecolor{arylideblue}{rgb}{0.8,0.9,1.0}
\definecolor{arylideyellow}{rgb}{0.98, 0.91, 0.71}
\definecolor{arylidepink}{rgb}{0.96, 0.76, 0.76}

\AtBeginDocument{%
  }

\copyrightyear{2022}
\acmYear{2022}
\setcopyright{acmcopyright}
\acmConference[ICMI '22]{INTERNATIONAL CONFERENCE ON MULTIMODAL INTERACTION}{November 7--11, 2022}{Bengaluru, India}
\acmBooktitle{INTERNATIONAL CONFERENCE ON MULTIMODAL INTERACTION (ICMI '22), November 7--11, 2022, Bengaluru, India}
\acmPrice{15.00}
\acmDOI{10.1145/3536221.3556624}
\acmISBN{978-1-4503-9390-4/22/11}

\begin{document}

\title{Multimodal Across Domains Gaze Target Detection}

\author{Francesco Tonini}
\affiliation{
    \department{Department of Information Engineering and Computer Science}
	\institution{Department of Information Engineering and Computer Science (DISI), University of Trento}
	\city{Trento}
	\country{Italy}
}
\email{francesco@tonini.dev}

\author{Cigdem Beyan}
\affiliation{
    \department{Dept. of Information Engineering and Computer Science}
	\institution{Department of Information Engineering and Computer Science (DISI), University of Trento}
	\city{Trento}
	\country{Italy}
}
\email{cigdem.beyan@unitn.it}

\author{Elisa Ricci}
\affiliation{
    \department{Department of Information Engineering and Computer Science}
	\institution{Department of Information Engineering and Computer Science (DISI), University of Trento}
	\city{Trento}
	\country{Italy}
}
\affiliation{
    %\department{}
	\institution{ Deep Visual Learning Research Group, Fondazione Bruno Kessler}
	\city{Trento}
	\country{Italy}
}
\email{e.ricci@unitn.it}

\renewcommand{\shortauthors}{Tonini et al.}
.
\begin{abstract}
This paper addresses the gaze target detection problem in single images captured from the third-person perspective. We present a multimodal deep architecture to infer where a person in a scene is looking. This spatial model is trained on the head images of the person-of-interest, scene and depth maps representing rich context information. Our model, unlike several prior art, do not require supervision of the gaze angles, do not rely on head orientation information and/or location of the eyes of person-of-interest.
Extensive experiments demonstrate the stronger performance of our method on multiple benchmark datasets. We also investigated several variations of our method by altering joint-learning of multimodal data. Some variations outperform a few prior art as well. First time in this paper, we inspect domain adaptation for gaze target detection, and we empower our multimodal network to effectively handle the domain gap across datasets. The code of the proposed method is available at \url{https://github.com/francescotonini/multimodal-across-domains-gaze-target-detection}.

\end{abstract}




\maketitle

\section{Introduction} \label{sec:intro}
Gaze behavior indicates the visual attention of a person and allows to specify what a person is interested in, helps to decipher and forecast the interactions, intentions or actions of people~\cite{thakur2021predicting,fang2021dual,Niewiadomski2018}. Human-beings have a remarkable capability to detect the gaze direction of others,
understand whether a person is gazing them, follow other's gaze to identify their target, and determine the attention of others~\cite{Chong_2018_ECCV}. However, automatically performing and quantifying these remains as a challenging problem. The research on automatic gaze behavior analysis is divided as \emph{gaze estimation} and \emph{gaze target detection}~\cite{Recasens2017,Zhengxi2022,chong2020detecting,fang2021dual}. Gaze estimation refers to determining the person's gaze direction (typically in 3D) while does not focus on accurately locating where a person in the scene is looking~\cite{YiLehWu2014,zhang2018training,Liu2020}. Instead, gaze target detection (also referred as \emph{gaze-following}~\cite{chong2020detecting,fang2021dual,jin2022depth}) is to inferring where each person in the scene (2D or 3D) is looking~\cite{Recasens2017,thakur2021predicting,Liu2020}. 
This paper addresses the \textbf{\textit{gaze target detection}} in \textbf{\textit{single images}} (i.e., in 2D), collected \textbf{\textit{in-the-wild}}, and captured from the \textbf{\textit{third-person perspective}}.
In this scope, earlier works present Convolutional Neural Network (CNN)-based architectures composed of two-pathways. While one path learns feature embeddings from the scene images, the other one models the head image belonging to the person whose gaze target is to be predicted~\cite{Recasens2015,Recasens2017,lian2018believe}.
Studies ~\cite{Recasens2015,Recasens2017,lian2018believe} perform spatial modeling, instead Chong et al.~\cite{chong2020detecting} extended the aforementioned two-pathway architecture by explicitly modeling the embeddings of the scene and head images over time (i.e., apply spatio-temporal modeling).
That method~\cite{chong2020detecting} presents improved results with respect to earlier research, however, it still lacks of understanding the so-called person-relative depth. As a consequence, false detections occur when there are multiple object-of-interests at different depths but along with the subject's gaze direction.
This handicap was handled in~\cite{fang2021dual,jin2022depth} by integrating the depth images into the pipeline. Fang et al.~\cite{fang2021dual} additionally relies on head pose detection, eyes detection and eye features extraction. Such a framework~\cite{fang2021dual} improved the gaze target detection performance, while potentially being error-prone in real-life processing, e.g., when the eyes are not visible or detectable.
On the other hand, Jin et al.~\cite{jin2022depth} involves an auxiliary network to perform 3D-gaze orientation estimation using pseudo labels, in addition to using another auxiliary network to estimate the depth. The performance of~\cite{jin2022depth} depends on reliable depth and orientation pseudo labels.

Unlike~\cite{Recasens2015,Recasens2017,lian2018believe}, we do not require supervision of gaze angles, which simplifies our training process and improves its applicability. Different from~\cite{chong2020detecting}, we apply only spatial processing, but still able to detect the gaze target at each frame of a video. Similar to~\cite{fang2021dual,jin2022depth}, we use depth images. Our \textbf{\textit{multimodal pipeline}} (see Sec.~\ref{subsec:multiNet}) has three-pathways to process: \textit{i)} the head image, \textit{ii)} the scene image and \textit{iii)} the depth map, which is obtained by standalone monocular depth estimation from RGB images~\cite{ranftl2020towards}.
It is important to highlight that our pipeline is computationally low-cost and simpler than~\cite{fang2021dual} by not requiring the detection of the head pose and the location of the eyes.
Furthermore, unlike~\cite{jin2022depth}, our proposal does not use an additional network to estimate gaze orientation from head features. Instead, we implicitly learn the orientation features using the head attention module. Given the proposed pipeline, one major aspect of this paper is to investigate how different modalities should be jointly learned for performing effective gaze target detection. To do so, we present a comprehensive experimental analysis (see Sec.~\ref{sec:ablation}). Consequently, not only the proposed method but also some of the variations of it exceed the performance of the prior art.

Generalization capability of a trained gaze target detection model is of paramount importance for its utilization in practice. However, empirical analysis (Sec.~\ref{sec:acrossDomains}) show that, the performance of a gaze target detection model significantly decreases when it is tested on a dataset different from the one it is trained on. This phenomena is equally valid when the training is performed on the in-the-wild datasets (Sec.~\ref{sec:datasets}).
Motivated by this, as the first attempt in the gaze target detection literature, this paper studies the domain-shift problem, and propose a novel \textbf{\textit{domain adaptation method}} integrated into the proposed multimodal gaze target detection architecture (Sec.~\ref{subsec:DA}). Our method improves the results remarkably by also outperforming a state-of-the-art (SOTA) domain adaptation method. The main contributions of this study can be summarized as follows.
\vspace{-0.2cm}
\begin{itemize} [wide, labelwidth=!, labelindent=0pt]
    \item A novel multimodal deep architecture, that detects the gaze target in a 2D-image captured from the third-person perspective, is proposed. 
    \item We empirically validate the performance of the proposed model on several benchmark datasets in which it shows improved results relative to the SOTA.
    \item We show the effectiveness of the proposed model through an ablation study and by presenting several variations of it. Even some of the variations achieve better scores compared to the SOTA.
    \item This work is the first attempt where the domain adaption for gaze target detection is studied. We first diagnosed the domain-shift problem for our model as well as the prior art, and then propose a novel method to handle it.
    \item The proposed domain adaptation approach results in enhanced performance on target datasets, which is also superior than a SOTA multimodal domain adaptation method.
\end{itemize}

\section{Related Work}\label{sec:relatedWork}
Below, we describe the related studies for gaze target detection task. Then, we briefly summarize the domain adaptation (DA) research in general, and focus on DA for gaze behavior analysis and multimodal visual data.

\subsection{Gaze Target Detection}\label{sec:gazetargetdet}
Gaze target detection has applications in several fields, e.g., human interaction systems, computer vision and robotics, where it is important to understand the object-of-interest~\cite{Schauerte2014}, predict and anticipate the actions~\cite{Min_2021_WACV,li2021eye} and so forth. Most existing works on gaze target detection rely on a particular sensor (eye trackers~\cite{thakur2021predicting}, VR/AR devices~\cite{Dohan2019}, RGB-D cameras~\cite{wei2018,Zhengxi2022}, etc.) or applicable for specialized settings (e.g., face-to-face meetings~\cite{Beyan2016}) or applications (e.g., identifying the mutual gaze~\cite{Marin-Jimenez_2019_CVPR}, detecting the common gaze point of multiple human observers~\cite{zhuang2019muggle,yang2021gaze}, anticipating averted gaze~\cite{muller2020anticipating})
or requires constrained subject placement in the scene~\cite{masse2019extended}. Another categorization is regarding whether the target is in 2D image~\cite{Recasens2015,Recasens2017,Chong_2018_ECCV,lian2018believe,chong2020detecting} or 3D space~\cite{masse2017tracking,wei2018,brau2018multiple,Zhengxi2022}.

In this paper, we focus on the gaze target detection in \emph{2D-single images} which are collected in \emph{unconstrained environments} from the \textit{third-person view}. In this context, one of the first work adapting deep learning architectures was~\cite{Recasens2015}, which present two-pathway architecture. One branch of that network~\cite{Recasens2015} takes the scene images to estimate the saliency (so-called saliency pathway) while the other one (so-called gaze pathway) gets head images as the input and models the gaze direction. An effective component of that network~\cite{Recasens2015} is the head location information injected into the gaze pathway, which improves the gaze target detection results remarkably. 
The posterior work~\cite{Chong_2018_ECCV} adapted the aforementioned two-pathway architecture while others~\cite{Recasens2017,lian2018believe,chong2020detecting,fang2021dual,jin2022depth} utilized both the two-pathway model and the head information injection pipeline. Differently, Chong et al.~\cite{Chong_2018_ECCV} extended the architecture of~\cite{Recasens2015} to detect the gaze targets not-being in the scene (so called out-of-frame gaze targets) by simultaneously learning the gaze angles and the saliency.
The out-of-frame component (\textit{two convolutional layer + ReLU + softmax}) is kept the same in ~\cite{chong2020detecting,fang2021dual}.
Chong et al.~\cite{chong2020detecting} additionally integrated CNN-LSTM, which processes the feature embeddings of the gaze and saliency pathways to learn the gaze behavior in time. 
Even though~\cite{Recasens2015,Recasens2017,lian2018believe,chong2020detecting} present promising results for gaze target detection, they all fail to address the challenge of handling the situations where the person-relative depth matters. For instance, in the situations where there are multiple objects at different depths along with the subject's gaze direction, it is unlikely that the correct gaze target can be determined by any of these methods. In order to handle this,~\cite{fang2021dual} utilize the depth information and 3D-gaze to produce target-focused spatial attention map. However, the overall pipeline of~\cite{fang2021dual} is computationally heavy as it requires detection of head pose and eyes. Jin et al.~\cite{jin2022depth} is another study which include the depth maps. The authors~\cite{jin2022depth} designed a primary network that predicts gaze as in~\cite{chong2020detecting}. Furthermore, to improve the prediction performance, they introduce two auxiliary networks: one to learn depth features, the other to learn 3D-gaze orientation features. The whole pipeline is learnt using the ground-truth, pseudo depth and orientation labels.

Our work diverges from prior art in several aspects. First of all, unlike~\cite{Recasens2015,Recasens2017,lian2018believe}, we do not require supervision of gaze angles. Different from~\cite{Recasens2015,Recasens2017,lian2018believe,chong2020detecting}, we adapt the depth images (obtained by monocular depth estimation method~\cite{ranftl2020towards}) and consequently improve the spatial modeling by handling the person-relative depth challenge. Unlike~\cite{chong2020detecting}, we not only apply spatial pooling in the scene network by using the head features that supplies a regulation through attention mechanism, but also integrate this for the depth network, resulting in improved performance. Also, we only rely on the spatial information, and do not apply spatio-temporal data processing (i.e., less training time with respect to~\cite{chong2020detecting}).
We present a three-pathway network and joint learning with late fusion of head location regulated, scene and depth feature embeddings, without using head poses and location of eyes as applied in~\cite{fang2021dual}. Unlike~\cite{jin2022depth} our work neither requires additional depth and orientation pseudo labels nor additional networks to explicitly learn a 3D orientation representation.

We adapted the out-of-frame component of~\cite{chong2020detecting}, but this is performed attached to the multimodal framework, which is not the case in~\cite{chong2020detecting,fang2021dual}. Our framework achieves better performance compared to prior art, and in some cases even surpasses the human performance. Importantly, this is the first work investigating the domain-shift problem for gaze target detection in 2D images, and presenting a relatively simple but effective multimodal DA method to boost the generalizability of the proposed three-pathway network.

\subsection{Domain Adaptation}\label{sec:domainAdapt}
Unsupervised Domain Adaptation (UDA) is a broadly investigated methodology in order to handle the problems coming out due to the domain gap, which can happen when the training and testing data are belonging to different distributions. UDA transfers knowledge from a labeled dataset (called source domain) to another domain (called target domain), whose data is available at training time but \emph{without labels}~\cite{10.1145/3400066}.
The literature of UDA can be divided into three as: \textit{i)} discrepancy-based techniques~\cite{10.1145/2663204.2663247,long2015learning,xu2019larger} that try to minimize the distance between source and target distributions at feature level, \textit{ii)} adversarial methods~\cite{8237506,cui2020gradually} having a generator and a discriminator and trying to have features created by the generator as close as possible to those of the source, and \textit{iii)} self-supervised methods~\cite{xu2019self,Wang_2021_ICCV,da_Costa_2022_WACV} optimizing the (self-supervised) objective function to produce robust representations for the main task.

There exist relatively few research addressing to adapt to unseen domains for gaze estimation task, while to the best of our knowledge, this has not yet been investigated for the gaze target detection in 2D images (i.e., the aim of this paper). For the gaze estimation problem, Kellnhofe et al.~\cite{kellnhofer2019gaze360} adapts the adversarial discriminative DA of~\cite{Tzeng_2017_CVPR} in which a discriminator identifies the source domain of the image features as a binary classification task in addition to having another loss used to exploit the left-right symmetry of the gaze-estimation task, providing the consistency on unlabeled data by computing the gaze of the original and horizontally flipped images in order to minimize the angular difference among two.
Yu et al.~\cite{yu2019improving} approach the gaze adaptation problem in terms of gaze redirection given that the eye structures of different persons cause a domain gap resulting in poor performance. To handle this, authors~\cite{yu2019improving} generate synthetic eye images from existing reference samples (i.e., self-supervised DA) and define the gaze redirection loss (calculated based on enforcing that the gaze predicted from gaze redirected image is close to its target ground-truth) in addition to cycle consistency loss of~\cite{8237506}. Recently, Guo et al.~\cite{guo2020domain} present UDA gaze estimation by embedding with prediction consistency, which ensures that linear relationships between gaze directions in different domains remain consistent on gaze space and embedding space.

On the other hand, for multimodal visual data DA, most of the prior art considers single modality. A major number of work investigated the domain-shift across RGB images while depth images were left behind. There exist methods~\cite{hoffman2016cross,Hoffman_2016_CVPR,spinello2012leveraging} defining RGB images as source, and depth images (obtained from RGB-D camera) as target. Xiao et al.~\cite{Xiao2017} defines the RGB-D as source while RGB as the target. Instead~\cite{wang2019unsupervised,bousmalis2017unsupervised} use the depth information as the additional channel for source and target.
Unlike aforementioned litreature, ~\cite{ferreri2021multi} recently presents a self-supervised modality translation, showing the SOTA results for RGB-D scene recognition. Ferreri et al.~\cite{ferreri2021multi} defines two encoder-decoder architecture based on ResNet-18; one for RGB branch and the other for depth branch. The depth decoder reconstructs the RGB image when the encoder's output is the depth embeddings, and the RGB decoder reconstructs the depth image when the encoder's output is the RGB embeddings. Authors also use an additional ResNet-18 to regulate the content similarity between the reconstructed images and the original images. 
In this paper, we adapt the method of ~\cite{ferreri2021multi} into our multimodal network to perform gaze target detection, whose results are compared with the proposed DA method.

The proposed DA method conducts in three networks: head, scene and depth, simultaneously. We inject a Gradient Reversal Layer (GRL) \cite{ganin2016domain} between our head backbone and a domain classifier we define (which decides whether a head image belongs to the source or the target domain). We also apply {RGB$\rightarrow$Depth} and {Depth$\rightarrow$RGB} modality translations by attaching additional decoders to our scene and depth backbones.

\begin{figure*}[t!]
    \centering
    \includegraphics[width=\textwidth]{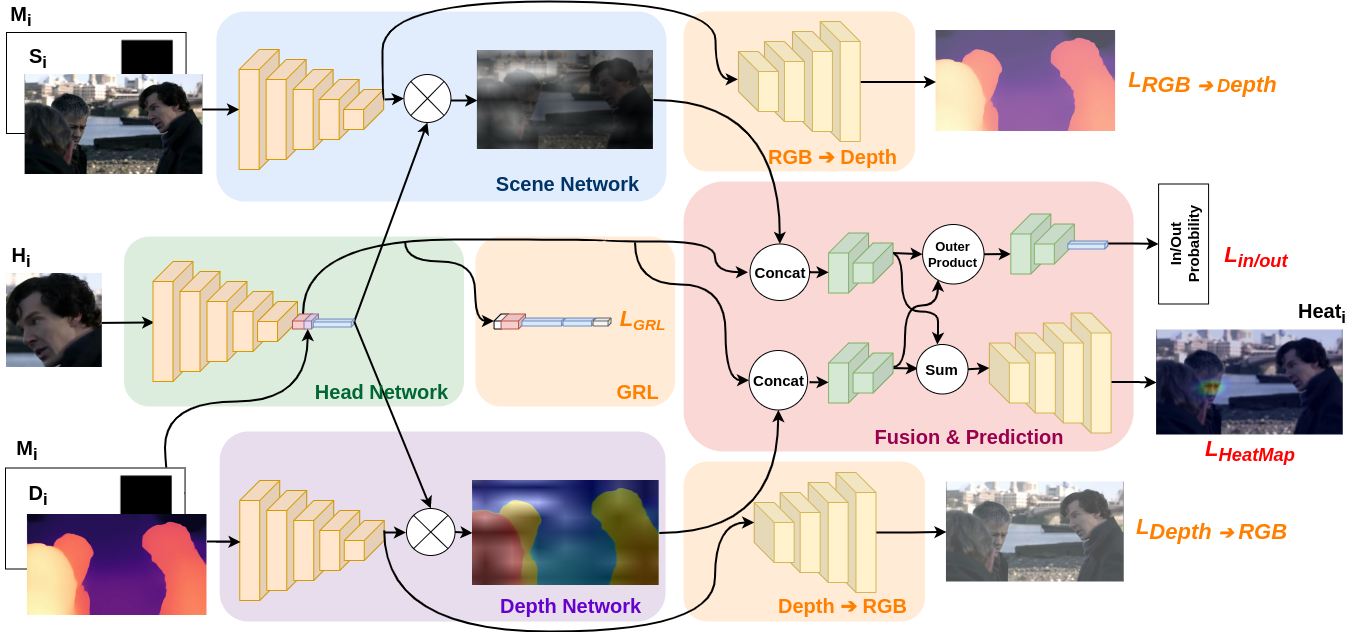}
    \caption{Our \textbf{three-pathway architecture} composed of \textbf{scene network} (blue box), \textbf{head network} (green box) and \textbf{depth network} (purple box). 
    The input of these networks, respectively, are: \textit{i)} scene image ($\mathbf{S}_i$) concatenated with head location mask ($\mathbf{M}_i$; a binary image, the bounding box of the person whose gaze to be predicted is black, the rest is white), \textit{ii)} head image ($\mathbf{H}_i$) belonging to the person whose gaze to be predicted, and \textit{iii)} depth map ($\mathbf{D}_i$) obtained by~\cite{ranftl2020towards} concatenated with $\mathbf{M}_i$. We also present the intermediate images obtained by scene and depth networks, where the effect of head attention can be seen.
    The multimodal joint learning and gaze prediction are performed by the \textbf{fusion and prediction component} (pink box). The outputs of our architecture are: \textit{iv)} 2D heatmap ($\mathbf{Heat}_i$, superimposed on the scene image, associated loos is L$_{HeatMap}$) and \textit{v)} the probability of the gaze being inside or outside the scene (loss shown as L$_{in/out}$).
    We also perform \textbf{domain adaptation} (yellow boxes) \textit{vi)} attached to head network (shown as \textbf{GRL}, with the loss: L$_{GRL}$), \textit{vii)} attached to depth network to perform adaptation from depth to RGB images (shown as \textbf{Depth$\rightarrow$RGB}, with loss: L$_{Depth \rightarrow RGB}$), and \textit{viii)} attached to scene network to perform adaptation from RGB to depth (shown as \textbf{RGB$\rightarrow$Depth}, with the loss: L$_{RGB \rightarrow Depth}$).
    }
    \Description{Two man and one woman are having conversation. The man on right gazes the other man on his left.}
    \label{fig:proposedM}
\end{figure*}

\section{Methodology}\label{sec:ProposedMethod}

Given an RGB image $\mathbf{S}_i$ of the scene, the depth map $\mathbf{D}_i$ obtained from $\mathbf{S}_i$ using the state-of-the-art (SOTA) monocular depth estimator~\cite{ranftl2020towards},
%Furthermore, let 
$\mathbf{H}_i$; a region of interest in $\mathbf{S}_i$ which contains the head of the person (i.e., the gaze source), and a binary head location mask $\mathbf{M}_i$ in the size of $\mathbf{S}_i$ when $M_i(x, y) = 1$ for each pixel of the person's head,
the aim of our architecture is to generate a 2D heatmap $\mathbf{Heat}_i$ in the size of $\mathbf{S}_i$ when the higher values are closer to the ground-truth gaze coordinate and $\underset{x, y}{\mathrm{argmax}} \; Heat_i(x, y)$ is the gaze target in pixel coordinates. In addition to $\mathbf{Heat}_i$, our neural network also has the output $\mathbf{InOut}_i$ such that $\mathbf{InOut}_i = 1$ if and only if the gaze target is inside the frame.

Our proposal builds on the work of Chong et al.~\cite{chong2020detecting} and injects the depth modality, which together with RGB modality supplies a richer representation of the scene, and consequently performs better in the challenging scenarios. 
The proposed network is composed of multiple self-contained modules. The \textit{\textbf{scene and depth networks}} ($\mathbf{SN}$ and $\mathbf{DN}$) process $\mathbf{S}_i$ and $\mathbf{D}_i$, respectively. The \textit{\textbf{head network}} ($\mathbf{HN}$) processes $\mathbf{H}_i$ independently, and produces an attention map that is then multiplied by the embeddings of $\mathbf{SN}$  and $\mathbf{HN}$.
The \textit{\textbf{fusion and prediction module}} ($\mathbf{F\&P}$) concatenates the final scene, depth, and head features to obtain the outputs of our proposal: $\mathbf{Heat}_i$ that encodes the region in which gaze happens, and $\mathbf{InOut}_i$; the probability of the gaze target being inside or outside the scene. 
Different from~\cite{chong2020detecting}, we also introduce the channel-wise outer product \cite{sun2020learning} and summation operators for multimodal joint learning. Furthermore, we bring in a \textbf{\textit{domain adaptation} (DA) module} that relies on a Gradient Reversal Layer~\cite{ganin2016domain} on $\mathbf{HN}$
as well as two modality decoders that work on $\mathbf{SN}$ and $\mathbf{DN}$, respectively.
An illustration of our network is given in Fig.~\ref{fig:proposedM}. The code is available at \url{https://github.com/francescotonini/multimodal-across-domains-gaze-target-detection}. Below, we split our proposal into two sections: Multimodal Network (Sec.~\ref{subsec:multiNet}) and Domain Adaption (Sec.~\ref{subsec:DA}), and describe each aforementioned module in detail. Lastly, Sec.~\ref{subsec:impDetails} supplies the implementation details.

\subsection{Multimodal Network}\label{subsec:multiNet}

\noindent
\textbf{Head Network.}
Given the RGB scene image $\mathbf{S}_i$, we crop the head of the person-of-interest to obtained head image $\mathbf{H}_i$.
$\mathbf{H}_i$ is processed by the head network's backbone that maps the original representation $\mathbf{H}_i$ into a feature embedding $\mathbf{e}_{i}^{H}$.
Such features are average pooled and processed by a set of linear layers that outputs an attention map.
The outcome of the attention map $\mathbf{attn}_{i}^{H}$ is multiplied by the scene and depth feature embeddings.

\noindent
\textbf{Scene Network.}
The scene network shares the backbone structure of the head and depth networks. This module takes as the input the concatenation of the RGB scene image $\mathbf{S}_i$ and the binary head location mask $\mathbf{M}_i$ that encodes the position of the person's head in the scene image.
Each channel of the feature embedding of the scene network $\mathbf{e}_{i}^{S}$ is multiplied by the attention map $\mathbf{attn}^{H}_{i}$ generated by the head network:
\begin{equation}
    \mathbf{e}_{i}^{S*} = \mathbf{e}_{i}^{S} \otimes \mathbf{attn}^{H}_{i}, 
\end{equation}
where $\otimes$ is the channel-wise multiplication.
By multiplying the output of the scene network's backbone with the attention map, we force the network to focus on objects in the scene that are relevant with respect to the person-of-interest and its head orientation. This is in line with SOTA~\cite{chong2020detecting,fang2021dual}.

\noindent
\textbf{Depth Network.}
The depth network shares the same backbone structure and the input shape of the scene network $\mathbf{SN}$.
This module takes as input the depth map $\mathbf{D}_{i}$ of the scene and the binary head location mask $\mathbf{M}_i$.
The feature embeddings $\mathbf{e}_{i}^{D}$ from the depth backbone are multiplied by the head attention map $\mathbf{attn}_{i}^{H}$:
\begin{equation}
    \mathbf{e}_{i}^{D*} = \mathbf{e}_{i}^{D} \otimes \mathbf{attn}^{H}_{i},  
\end{equation}
where $\otimes$ is the channel-wise multiplication.

\noindent
\textbf{Fusion \& Prediction Network.}
The feature embeddings from the head network $\mathbf{e}_{i}^{H}$, the \textit{attended} scene $\mathbf{e}_{i}^{S*}$ and the \textit{attended} depth embeddings $\mathbf{e}_{i}^{D*}$ are the inputs of the fusion and prediction network $\mathbf{F}$.
The fusion module contains two encoder $\mathbf{ES}$ and $\mathbf{ED}$ that process the concatenation of $\mathbf{e}_{i}^{H}$, $\mathbf{e}_{i}^{S*}$, and $\mathbf{e}_{i}^{H}$, $\mathbf{e}_{i}^{D*}$ independently:
\begin{equation}
    \mathbf{e}_{i}^{HS*} = \mathbf{ES}(concat(\mathbf{e}_{i}^{H}, \mathbf{e}_{i}^{S*})),
\end{equation}
\begin{equation}
    \mathbf{e}_{i}^{HD*} = \mathbf{ED}(concat(\mathbf{e}_{i}^{H}, \mathbf{e}_{i}^{D*})),
\end{equation}
where $concat(\mathbf{A}, \mathbf{B})$ is the channel-wise concatenation of the feature embeddings of $\mathbf{A}$ and $\mathbf{B}$.

The prediction module outputs the 2D gaze heatmap $\mathbf{Heat}_i$ and the in/out of frame probability $\mathbf{InOut}_i$.
To obtain the 2D gaze heatmap $\mathbf{Heat}_i$, this module uses a multi-layer decoder $\mathbf{D}$ that takes as input the channel-wise summation $\oplus$ of scene and depth embeddings:
\begin{equation} \label{eq:head_i}
    \mathbf{Heat}_{i} = \mathbf{D}(\mathbf{e}_{i}^{HS*} \oplus \mathbf{e}_{i}^{HD*}),
\end{equation}
Furthermore, the channel-wise outer product \cite{sun2020learning} between scene and depth embeddings is input to a smaller encoder $\mathbf{EInOut}$ that produces the $\mathbf{InOut}_i$:
\begin{equation}
    \mathbf{InOut}_{i} = \mathbf{EInOut}(outer(\mathbf{e}_{i}^{HS*}, \mathbf{e}_{i}^{HD*})),
\end{equation}

\subsection{Domain Adaptation}\label{subsec:DA}
The multimodal network is enriched by multiple domain adaptation (DA) components that improve the performance of our proposed method when it is trained and tested across datasets. More specifically, our DA components attempt to improve the performance of our three networks: head, scene, and depth.

First, we introduce a domain classifier that performs a binary classification between the source and target domain represented in terms of the embeddings of the head backbone $\mathbf{e}_{i}^{H}$.
The domain classifier is connected to head backbone via Gradient Reversal Layer (GRL)~\cite{ganin2016domain}, which multiplies
the gradient by a certain negative constant during the backpropagation-based training. In other words, the domain classification loss is minimized for both source and target samples and the GRL ensures that the learned features are as indistinguishable as possible~\cite{ganin2016domain} (we simply show this as L$_{GRL}$ in Fig.~\ref{fig:proposedM}). The choice of adapting GRL is due to the fact that it has been one of the most popular UDA method, adapted to handle domain-shift problem for several applications, e.g., intention detection~\cite{zunino2020predicting}, view-invariant action recognition~\cite{UHAR_BMVC2021}, object recognition~\cite{tzeng2015simultaneous} and re-identification~\cite{ganin2016domain}.

Second, two additional decoders are integrated to the proposed method to reconstruct the original input $\mathbf{S}_i$ using the embedding of the depth network $\mathbf{e}_{i}^{D}$, and vice-versa. Thus, we perform a modality translation from RGB to depth, and depth to RGB. The implementation details clarify the associated loss functions.

\subsection{Implementation Details} \label{subsec:impDetails}
We implemented our model in PyTorch. The input to scene, depth, and head networks are normalized and resized to $224 \times 224$.
The backbones are based on ResNet-50 with an additional layer that creates a feature embedding of 1024 channels of size $7 \times 7$ in line with~\cite{Chong_2018_ECCV,chong2020detecting}.
The scene backbone is pre-trained on Places dataset~\cite{zhou2014learning}, and the head backbone is pre-trained on Eyediap dataset~\cite{funes2014eyediap} as applied in~\cite{Chong_2018_ECCV,chong2020detecting,fang2021dual,jin2022depth}, which show improved performance compared to using Vanilla ResNet-50. 
The depth backbone is also pre-trained on Places dataset~\cite{zhou2014learning}, after obtaining magma-colored depth maps by applying~\cite{ranftl2020towards} for each RGB image in the training set of~\cite{zhou2014learning}.

To perform fair analysis with the prior art~\cite{Recasens2015,Recasens2017,lian2018believe,Chong_2018_ECCV,chong2020detecting,fang2021dual}, the ground-truth gaze heatmap is obtained by plotting a Gaussian distribution around the center of gaze, i.e., the ground-truth gaze coordinate with respect to the scene image.
The total loss of \textbf{our multimodal network} is a weighted sum of the Mean Squared Error ($MSE$) loss on the gaze heatmap $L_{heatmap}$ and a binary cross entropy ($BCE$) loss $L_{in/out}$ for the in/out output:
\begin{equation}
    L_{total} = w_{heatmap} \; L_{heatmap} + w_{in/out} \; L_{in/out},
\end{equation}
where $w_{heatmap}$ and $w_{in/out}$ are the learnable weights.

Our multimodal network was trained from scratch on GazeFollow~\cite{Recasens2015} for 70 epochs with the batch size of 16 and the learning rate of $2.5 \times 10^{-4}$.
Afterwards, we fine-tuned the model on VideoAttentionTarget~\cite{chong2020detecting} following the implementation of SOTA~\cite{chong2020detecting,fang2021dual,tomas2021goo}.
Furthermore, we trained our multimodal network from scratch on GOO~\cite{tomas2021goo} for 70 epochs with the batch size of 16 and the learning rate of $2.5 \times 10^{-4}$.

To perform proposed \textbf{domain adaptation} method, at each step of the training, we forward a batch from the source domain followed by a batch from the target domain. The training was for up to 70 epochs, with a batch size of 16 and the learning rate of $2.5 \times 10^{-4}$. Due to the absence of the \textit{in/out} annotation on multiple datasets (see Sec.~\ref{sec:conclusion} for more details), while applying DA, we do not minimize the cross entropy loss: $L_{in/out}$.
The total loss ($L_{total_{w/DA}}$) while applying our DA module includes three additional losses: the cross entropy loss on the head domain classifier ($L_{GRL}$), the $MSE$ reconstruction losses from RGB to Depth ($L_{RGB \rightarrow Depth}$) and Depth to RGB ($L_{Depth \rightarrow RGB}$), shown as:
\begin{equation}
  \begin{aligned}
    L_{total_{w/DA}} = w_{heatmap} \; L_{heatmap} 
    + w_{GRL} \; L_{GRL}
    + w_{RGB \rightarrow Depth} \; L_{RGB \rightarrow Depth} 
    + w_{Depth \rightarrow RGB} \; L_{Depth \rightarrow RGB},
  \end{aligned}
  \label{eq:daFull}
\end{equation}
where $w_{heatmap}$, $w_{GRL}$, $w_{RGB \rightarrow Depth}$, and $w_{Depth \rightarrow RGB}$ are the learnable weights.

\section{Experimental Analysis} 
\label{sec:results}
We conducted a comprehensive analysis to evaluate the performance of our method. Sec.~\ref{sec:ablation} presents an ablation study for our multimodal network to show the contribution of the scene, depth and head networks. It also includes an extensive investigation regarding modality fusion. Sec.~\ref{sec:compSOTA} compares the performance of the proposed multimodal network against to the prior art. In Sec.~\ref{sec:acrossDomains}, we study gaze target detection task across datasets. The experiments given in Sec.~\ref{sec:ablation}-~\ref{sec:acrossDomains} do not apply any DA method, while the experiments in Sec.~\ref{sec:DA} corresponds to applying DA.
Our method (with / without DA) achieves the state-of-the-art results on all datasets in all experiments. Finally, we present the qualitative results of the proposed method (with / without DA) in Sec.~\ref{sec:qual}.

\subsection{Datasets \& Evaluation Metrics}
\label{sec:datasets}
The proposed method is evaluated on three benchmark datasets: GazeFollow~\cite{Recasens2015}, VideoAttentionTarget~\cite{chong2020detecting} and Gaze On Objects (GOO)~\cite{tomas2021goo}. We follow the standard training and test splits of each dataset to supply fair comparisons with SOTA. \\

\noindent
\textbf{Datasets.} \textit{GazeFollow}~\cite{Recasens2015} dataset includes more than 120K images from various classification and detection datasets (i.e., SUN~\cite{xiao2010sun}, COCO~\cite{lin2015microsoft}, Actions-40~\cite{yao2011human}, PASCAL~\cite{pascal}, and Places~\cite{zhou2014learning}), with more than 130K annotations of head locations and the corresponding gaze points.
\textit{VideoAttentionTarget}~\cite{chong2020detecting} is a collection of 1331 video clips from various sources on YouTube. The annotations include more than 160K frame-level head bounding boxes and 110K gaze targets inside the scene.
\textit{Gaze On Objects (GOO)}~\cite{tomas2021goo} dataset is a collection of images of shelves with 24 classes of groceries. In each image, a person looks at one object on a shelf. Objects in the scene are annotated with their bounding box and class. GOO is the first dataset in the gaze target detection task that uses both real and synthetic data. Out of the 200K images of the datasets, 8K are captured from a real environment, while 192K are generated using a 3D engine that reconstructs the real environment. In this paper, we use the images belonging to real environment. \\

\noindent
\textbf{Evaluation Metrics.} The following metrics were adopted to evaluate the performance of the proposed model in line with the SOTA~\cite{Recasens2015,Chong_2018_ECCV,lian2018believe,chong2020detecting,fang2021dual}. \textit{Heatmap Area Under Curve (AUC \%)}~\cite{Judd2009} is to asses the confidence of the predicted heatmap with respect to the ground-truth. \textit{Average distance (Avg.Dist.)} stands for the Euclidean distance between the predicted gaze location and the ground-truth gaze point.

\subsection{Ablation Study \& Modality Fusion} \label{sec:ablation}
To better investigate the contribution of different components of our model (i.e., scene, head and depth networks) and to compare the effectiveness of the different modality fusion techniques, we trained the following variations. \textbf{1) Scene network only:} We remove the head network $\mathbf{HN}$ and the depth network $\mathbf{DN}$. The head features concatenated with the scene features are not provided, thus the only way to make the attention map $\mathbf{attn}_{i}^{H}$ is through the head location mask $\mathbf{M}_i$. \textbf{2) Scene and head networks:} This stands for removing the depth network $\mathbf{DN}$ while the scene $\mathbf{SN}$ and the head networks $\mathbf{HN}$ remain. The input of the scene network is the concatenation of the scene image $\mathbf{S}_i$ and the head location mask $\mathbf{M}_i$. \textbf{3) Grayscale depth and head networks:} This refers to removing the scene network $\mathbf{SN}$ while keeping the head $\mathbf{HN}$ and the depth networks $\mathbf{DN}$. The input of the depth network is the concatenation of the \textit{grayscale} depth map $\mathbf{D}_i$ and the head location mask $\mathbf{M}_i$. \textbf{4) Colored depth and head networks:} This is a similar set up with (3) when the depth map $\mathbf{D}_i$ is colored by magma-colormap (as shown in Fig.~\ref{fig:proposedM}). \textbf{5) Early fusion V1:} We replace the head location mask $\mathbf{M}_i$ with the \textit{grayscale} depth map $\mathbf{D}_i$ while the depth network $\mathbf{DN}$ is completely removed. In other words, this setup includes head network $\mathbf{HN}$ and the scene network $\mathbf{SN}$ when the input tensor of the scene network has four channels (three for scene image $\mathbf{S}_i$ and one for \textit{grayscale} depth map $\mathbf{D}_i$). \textbf{6) Early fusion V2:} This refers to removing the depth network $\mathbf{DN}$ and feeding $\mathbf{SN}$ with the input tensors having five channels (three for scene image $\mathbf{S}_i$, one for head location mask $\mathbf{M}_i$, and one for \textit{grayscale} depth map $\mathbf{D}_i$). \textbf{7) Early fusion V3:} This is a similar set up to (6) when the input tensors have seven channels (three for scene image $\mathbf{S}_i$, one for head location mask $\mathbf{M}_i$ and three for colored depth map $\mathbf{D}_i$). \textbf{8) Early fusion V4:} This refers to having a single encoder network instead of having $\mathbf{ES}$ and $\mathbf{ED}$ whose inputs are the concatenation of the scene feature map, depth feature map and the head feature map. \textbf{9) Depth-Aware Scene Convolutional Network (Early Fusion V5):} We remove the proposed $\mathbf{DN}$ and replace the $\mathbf{SN}$ with the depth-aware scene convolutional network of~\cite{wang2018depth} when the inputs of it are the grayscale depth map and the RGB scene images in parallel. \textbf{10) Late fusion with concatenation:} We replace the proposed summation operation in Eq.~\ref{eq:head_i} with concatenation, which is applied before the multi-layer decoder $\mathbf{D}$. \textbf{11) Late fusion with outer product:} It refers to replacing the applied summation operation in Eq.~\ref{eq:head_i} with the channel-wise outer product proposed in~\cite{sun2020learning}. These aforementioned variations were tested on the GazeFollow~\cite{Recasens2015} dataset. The most competitive ones were also validated on the VideoAttentionTarget~\cite{chong2020detecting} dataset. The corresponding results are given in Table~\ref{table:ablationStudy1}.

\begin{table}[t]
    \caption{Results of ablation study and the modality fusion on the GazeFollow~\cite{Recasens2015} and VideoAttentionTarget~\cite{chong2020detecting} datasets. Exp. follows the order of variations (1-11) described in Sec.\ref{sec:ablation}. For ease of reading, the ablation study, early fusion and late fusion are given in \colorbox{arylideyellow}{yellow}, \colorbox{arylidepink}{pink}, and \colorbox{arylideblue}{blue}, respectively. The best results (the higher the AUC and the lower the Avg.Dist.) are shown in bold.}
    \centering
 %   \resizebox{\linewidth}{!}
 %  {
    \begin{tabular}[t]{lcccccccccccc}  \hline \hline
               
                {Exp.} & \colorbox{arylideyellow}{1} & \colorbox{arylideyellow}{2} & \colorbox{arylideyellow}{3} & \colorbox{arylideyellow}{4} & \colorbox{arylidepink}{5} & \colorbox{arylidepink}{6} & \colorbox{arylidepink}{7} & \colorbox{arylidepink}{8} & \colorbox{arylidepink}{9} & \colorbox{arylideblue}{10} & \colorbox{arylideblue}{11} & \textbf{Ours} \\

                 \multicolumn{13}{c}{\bf GazeFollow~\cite{Recasens2015}} \\
                 
                {AUC (\%)} & 75.8 & 92.1 & 87.5 & 89.3 & 91.2 & 92.4 & 91.5 & 91.3 & 90.0 & 92.6 & 92.6 & \textbf{92.7} \\
                {Avg.Dist.} & 0.258 & 0.143 & 0.215 & 0.166 & 0.157 & 0.149 & 0.151 & 0.143 & 0.183 & 0.143 & 0.142 & \textbf{0.141} \\ \hdashline

              \multicolumn{13}{c}{\bf VideoAttentionTarget~\cite{chong2020detecting}} \\

              {AUC (\%)} & - & 90.6 & - & - & - & 93.0 & 93.3 & 93.8 & - & 90.5 & 90.6 & \textbf{94.0} \\
              Avg.Dist. & - & 0.139 & - & - & - & \textbf{0.129} & 0.139 & 0.132 & - & 0.143 & 0.141 & \textbf{0.129} \\
                \hline \hline

    \end{tabular}
   %     }
    \label{table:ablationStudy1}
\end{table}

%%%%%%%%%%%%% network components ablation starts here %%%%%%%%%%%%%
Overall, the results show that all components of the proposed multimodal network are important to achieve the best performance. The experiments 1, 2, 3 and 4 allow us to understand the contribution of the head, depth and the scene networks, respectively. Out of three, the most crucial component is the head network (improving the AUC by 16.9\% and decreasing the Avg.Dist. by 0.117 in GazeFollow dataset~\cite{Recasens2015}), which gives intuition regarding the head orientation of a person in the scene, and allowing scene and the depth networks to pay more attention to the features that are more likely to be attended to. The second most contributing component is the scene network (performs up to +4.6\% AUC and -0.072 Avg.Dist., compared to depth+head network in GazeFollow dataset~\cite{Recasens2015}). Still the usage of scene network without the depth network fails to detect the target in a different depth level than the person who is gazing and, indeed scene+depth+head (proposed method) achieves +0.6\% AUC and -0.002 Avg.Dist. compared to scene+head in GazeFollow dataset~\cite{Recasens2015} while the performance improvement of the proposed method is higher in  VideoAttentionTarget~\cite{chong2020detecting} (+3.4\% AUC and -0.01 Avg.Dist.). 
On the other hand, when the depth and head networks remain and the scene network is removed, (experiments 3 and 4), one can be in favor of using the colored depth map instead of grayscale depth map (+1.8\% AUC and -0.05 Avg.Dist. in GazeFollow dataset~\cite{Recasens2015}).
%%%%%%%%%%%%% network components ablation ends here %%%%%%%%%%%%%

%%%%%%%%%%%%% Early fusion starts here %%%%%%%%%%%%%
The examination regarding how to combine the modalities is composed of applying various early fusion and late fusion experiments. The early fusion experiments (Exp. 5, 6 and 7) combine the modalities in the input space. %such that the input tensor has 4 (3 for scene, 1 for grayscale depth), 5 (3 for scene, 1 for grayscale depth and 1 for head location mask) and 7 (3 for scene, 3 for colored depth map and 1 for head location mask) channels, respectively.
Discarding head location mask (Exp. 5) decreases the performance compared to Exp. 6 and 7 by up to -1.2\% AUC and +0.008 Avg.Dist. in GazeFollow dataset~\cite{Recasens2015}. In early fusion setups, using grayscale depth map surpasses the colored version by +0.9\% AUC and -0.002 Avg.Dist. in GazeFollow dataset~\cite{Recasens2015}. However, this trend was not observed for VideoAttentionTarget~\cite{chong2020detecting}, in which using colored depth map achieved slightly better results than grayscale depth map in terms of AUC (+0.3\%). It is important to notice that these experiments are relatively lightweight as having only scene convolutional network compared to the proposed method (late fusion) and Exp. 8. Exp. 8 differs from Exp. 5, 6 and 7 as it combines the embeddings of the modalities before being encoded. It performs worse than other early fusion variations by up to -1.1\% AUC in GazeFollow dataset~\cite{Recasens2015}, instead performed better than others by up to +0.8\% AUC and -0.007 Avg.Dist. in VideoAttentionTarget~\cite{chong2020detecting}.
We also adapted the recent work Depth-Aware Scene Convolutional Network~\cite{wang2018depth} by replacing it with the scene convolutional network. However, this lowered the results up to -2.4\% for AUC and +0.04 for Avg.Dist. compared to other early fusion applications.
%%%%%%%%%%%%% Early fusion ends here %%%%%%%%%%%%%

%%%%%%%%%%%%% Late fusion starts here %%%%%%%%%%%%%
We further investigate different ways of applying late fusion. In GazeFollow~\cite{Recasens2015} dataset, we observed that different late fusion operations (i.e., concatenation, channel-wise outer product~\cite{sun2020learning} and summation in Eq.~\ref{eq:head_i}) perform on par while the proposed summation operation achieves slightly better AUC (+0.1\%) than others. It is important to notice that, for that dataset, all early fusion methods (Exp. 5-9) achieve worse results compared to the all late fusion methods (Exp. 10-11) with the drop in the margin of 0.2-3.3\% for AUC and the increase in the margin of 0.007-0.024 for Avg.Dist. On the other hand, the proposed late fusion (see Eq.~\ref{eq:head_i}) achieves remarkable results on VideoAttentionTarget~\cite{chong2020detecting} compared to other late fusion methods. That is up to +3.5\% for AUC and -0.014 for Avg.Dist. This also presents that the effectiveness of the proposed method generalizes better across different datasets compared to the other early and late fusion approaches tested.

%%%%%%%%%%%%%%%%%%%%%%%%%%%%%%%%%%%%%%%%%%%%%%%%%%%%%%%
\subsection{Comparisons with the State-of-the-art} \label{sec:compSOTA}
We compare our multimodal network with several SOTA in Table~\ref{table:SOTA1}. These comparisons include the standard gaze analysis baselines, namely: \textit{i)} random, \textit{ii)} center bias, and \textit{iii)} fixed bias, whose results are taken from \cite{Recasens2015}. Random stands for generating a heatmap per pixel by sampling the values from a Gaussian distribution. In center bias, the prediction is always the center of the image. In fixed bias, the location of the prediction is in terms of the average of fixations from the training set for the heads located to a similar area with the test image.  

Our method achieves better results compared to all counterparts, and becomes SOTA for all datasets in terms of AUC. It surpasses even the human performance in GazeFollow~\cite{Recasens2015} (+0.3\% AUC) and VideoAttentionTarget~\cite{chong2020detecting} (+1.9\% AUC) datasets.
In particular, its relative performance improvements in VideoAttentionTarget~\cite{chong2020detecting} and GOO~\cite{tomas2021goo} datasets are obtrusive (3.5-11\% and 1.8-12.2\% AUC, respectively).
In terms of Avg.Dist., our method falls behind Fang et al.~\cite{fang2021dual} and Jin et al.~\cite{jin2022depth} while performing better than others. It is important to notice that Fang et al.~\cite{fang2021dual} presents more complex and less lightweight model compared to ours by having additional components to \textit{i)} extract the head pose, \textit{ii)} detect the eyes and \textit{iii)} extract the eye features, which might be infeasible to perform correctly in real-life application. On the other hand, we can argue that the auxilary networks used in Jin et al.~\cite{jin2022depth} for gathering the 3D-gaze orientation information and depth map help to improve Avg.Dist., while that method performs poorer than ours and many other SOTA in terms of AUC.

Moreover, one can observe that, even though being a spatial model, our method outperforms Chong et al. ~\cite{chong2020detecting}, which includes Convolutional LSTM network. Consequently, we can draw a conclusion that integrating the depth map generated from the RGB scene images, i.e., a multimodal approach such as ours, Fang et al.~\cite{fang2021dual} or Jin et al.~\cite{jin2022depth}, results in better gaze target detection performance compared to relying on RGB videos, i.e., performing spatio-temporal data processing as in~\cite{chong2020detecting}. The late fusion results obtained by slightly tuning the proposed method (see Table~\ref{table:ablationStudy1}, Exp. 10 and 11) also confirm this conclusion by outperforming all the prior art. Besides, it is important to notice that some of the variations presented in Sec.~\ref{sec:ablation}, while being less effective than the proposed method, are still able to surpass the existing methods \cite{Recasens2015,Chong_2018_ECCV,lian2018believe,chong2020detecting,fang2021dual,jin2022depth} particularly when tested on VideoAttentionTarget~\cite{chong2020detecting} dataset.

\begin{table}[t!]
    \caption{Evaluation on benchmark datasets. The best results, the higher the $AUC$ and the lower the average distance ($Avg.Dist.$) is better, are shown in bold. $\star$ indicates our training. $\diamond$ taken from~\cite{tomas2021goo}. $Ours$ refers to the proposed multimodal network without domain adaptation. See text for the description of Random, Center and Fixed Bias.}
    \centering
 %   \resizebox{0.7\linewidth}{!}
%    {
    %{\begin{tabular}{p{0.5cm}p{0.3cm}p{0.7cm}|p{0.3cm}p{0.7cm}|p{0.3cm}p{0.7cm}} \hline \hline 
    
    {\begin{tabular}{lcc|cc|cc} \hline \hline 
    
        & \multicolumn{2}{c}{\bf \underline{GazeFollow~\cite{Recasens2015}}} &         \multicolumn{2}{c}{\bf \underline{VideoAttentionTarget~\cite{chong2020detecting}}} & \multicolumn{2}{c}{\bf \underline{GOO~\cite{tomas2021goo}}} \\ 
         & \multicolumn{1}{c}{{AUC}} & \multicolumn{1}{c}{{Avg.Dist.}} & \multicolumn{1}{c}{{AUC}} & \multicolumn{1}{c}{{Avg.Dist.}} & \multicolumn{1}{c}{{AUC}} & \multicolumn{1}{c}{{Avg.Dist.}}  \\  \hline
        Random & 50.4 & 0.484  & 50.5 & 0.458 & - & - \\
        Center & 63.3 & 0.313  & - & - & - & - \\
        Fixed Bias & 67.4 & 0.306  & 72.8  & 0.326 & - & - \\
        Recasens et al. \small{(2015)}~\normalsize{\cite{Recasens2015}} & 87.8 & 0.190  & - & - & 85.0$^\diamond$ & 0.220$^\diamond$ \\
        Chong et al. \small{(2018)}~\normalsize{\cite{Chong_2018_ECCV}} & 89.6 & 0.187  & 83.0 & 0.193 & - & - \\
        Lian et al. \small{(2018)}~\normalsize{\cite{lian2018believe}} & 90.6 & 0.145  & - & - & 84.0$^\diamond$ & 0.321$^\diamond$ \\
        Chong et al. \small{(2020)}~\normalsize{\cite{chong2020detecting}} & 92.1 & 0.137  & 86.0 & 0.134 & 79.6$^\diamond$ & 0.252$^\diamond$ \\
        Chong et al. \small{(2020)}~\normalsize{\cite{chong2020detecting}}$^\star$ & 92.2 & 0.143  & 86.2 & 0.136 & 90.0 & 0.190 \\
        Fang et al. \small{(2021)}~\normalsize{\cite{fang2021dual}} & 92.2 & 0.124  & 90.5 & \textbf{0.108} & - & - \\ 
        Jin et al. \small{(2022)}~\normalsize{\cite{jin2022depth}} & 92.0 & \textbf{0.118}  & 90.1 & 0.116 & - & - \\ 
        \textbf{Ours}  & \textbf{92.7} & 0.141 & \textbf{94.0} & 0.129 & \textbf{91.8} & \textbf{0.164} \\ \hline 
       \small{Human Performance} & 92.4 & 0.096  & 92.1 & 0.051 & - & - \\ \hline \hline
       
    \end{tabular}}
  %   }
  
    \label{table:SOTA1}
\end{table}

\subsection{Gaze Target Detection Across Datasets} \label{sec:acrossDomains}
This section examines the domain-shift problem for gaze target detection task in images and videos. To do so, we trained the model of Chong et al.~\cite{chong2020detecting}, and the proposed method on one dataset while the trained models were tested on a completely different dataset. The corresponding results are given in Table~\ref{table:acrossDataset}. As seen, our method outperforms Chong et al.~\cite{chong2020detecting} in all cross-domain analysis.
However, it is important to notice that both method (even though they were trained on in-the-wild datasets) suffer a lot (up to -36.5\% AUC and +0.3\% Avg.Dist.) when they were trained and tested on different domains with respect to training and testing them on the same domain. Consequently, we can argue that addressing domain-shift problem for gaze target detection task is inevitable.

\begin{table}[t]
    \caption{Evaluation of gaze target detection performance across datasets. The drop with respect to the same-dataset evaluation are given in parenthesis. Results in black are better than its counterpart. $\star$ indicates our training.}
    \centering
 %   \resizebox{0.70\linewidth}{!}
 %  {
    {\begin{tabular}[t!]{lcccc} \hline  \hline
    
        & Trained on & Tested on & AUC (\%) & {Avg.Dist.} \\ \hline 
       Chong et al. \small{(2020)}~\normalsize{\cite{chong2020detecting}}$^\star$ & GazeFollow & GOO & 77.3 (13.3$\downarrow$) & \textbf{0.270} (0.1$\downarrow$) \\
        \textbf{Ours} & GazeFollow & GOO & \textbf{78.3} (13.5$\downarrow$) & 0.284 (0.1$\downarrow$) \\ \hdashline
        
       Chong et al. \small{(2020)}~\normalsize{\cite{chong2020detecting}}$^\star$ & VideoAttentionTarget & GOO & 68.2 (22.4$\downarrow$) & 0.311  (0.1$\downarrow$) \\
        \textbf{Ours} & VideoAttentionTarget & GOO & \textbf{69.1} (22.7$\downarrow$) & \textbf{0.274} (0.1$\downarrow$) \\ \hdashline
        
       Chong et al. \small{(2020)}~\normalsize{\cite{chong2020detecting}}$^\star$ & GOO & GazeFollow & 62.5 (29.6$\downarrow$) & 0.410 (0.3$\downarrow$) \\
        \textbf{Ours} & GOO & GazeFollow & \textbf{62.9} (29.8$\downarrow$) & \textbf{0.401} (0.3$\downarrow$) \\ \hdashline
        
       Chong et al. \small{(2020)}~\normalsize{\cite{chong2020detecting}}$^\star$ & GOO & VideoAttentionTarget & 55.1 (30.9$\downarrow$) & 0.458 (0.3$\downarrow$) \\
        \textbf{Ours} & GOO & VideoAttentionTarget & \textbf{57.5} (36.5$\downarrow$) & \textbf{0.446} (0.3$\downarrow$) \\ \hline \hline

    \end{tabular}}
 %   }
    \label{table:acrossDataset}
\end{table}

\subsection{Domain Adaptation Results} \label{sec:DA}
Table~\ref{table:domanadapt} shows the results of the proposed domain adaption method (see Sec.~\ref{subsec:DA}). We evaluate our results as compared to Table~\ref{table:acrossDataset}. Additionally, we adapted the SOTA domain adaptation method Ferreri et al.~\cite{ferreri2021multi} to make comparisons among it's performance and ours. The corresponding results as well as an ablation study for the proposed DA module, are involved into the Table~\ref{table:domanadapt}.

Ferreri et al. (2021)~\cite{ferreri2021multi} is a SOTA for multimodal RGB-D scene recognition task, which takes as the inputs scene images in the format of RGB and depth. That pipeline ~\cite{ferreri2021multi} matches with our multimodal network. 
Consequently, we adapt Ferreri et al.~\cite{ferreri2021multi} into our multimodal network, also noticing that there is no DA method proposed for gaze target detection that we can conduct a comparison. To do so, we added two additional decoders to the proposed multimodal network to perform modality translation from RGB to depth, and depth to RGB. These decoders attempt to reconstruct the original input (e.g., scene image) using features extracted from the other modality (e.g., depth embeddings).
Such images were then compared against to the original input. An additional frozen ResNet-18 was also integrated to the proposed multimodal network to perform content similarity loss between reconstructed images and the original images. Such loss is obtained by calculating the $L1$ loss of multiple layers of the additional ResNet-18 network \cite{ferreri2021multi}. 
We trained our multimodal network using the code of~\cite{ferreri2021multi} with a learning rate of $2.5 \times 10^{-4}$ and for up to 40 epochs. We noticed that the test performances in earlier epochs (e.g., 3, 5) are better in all settings. We report the best of all test results of Ferreri et al.~~\cite{ferreri2021multi} in Table~\ref{table:domanadapt}.
Our DA pipeline differs from Ferreri et al.~\cite{ferreri2021multi} in terms of the following aspects: \textit{i)} instead of content similarity loss, we rely on a reconstruction loss between reconstructed and original images, and \textit{ii)} we also use a smaller decoder that shares the same structure of the \textit{fusion \& prediction network}'s decoder.
Precisely, it is composed of four \textit{convolutional + ReLU} blocks that reconstruct the RGB / Depth image starting from embeddings of Depth / RGB backbones. Compared to Ferreri et al.~\cite{ferreri2021multi}, one can notice that we present a much simpler and much lightweight DA component, which performs between \emph{scene} and \emph{depth} images. Additionally, integrating GRL further powers up the \emph{head features} consistency across source and target datasets. Indeed, the ablation study in Table~\ref{table:domanadapt} (applied when the source dataset is VideoAttentionTarget and target dataset is GOO) proves that each loss function of our DA module is important, and using them altogether results in the best performance.

\begin{table}[h!]
    \caption{Evaluation of domain adaptation methods for gaze target detection. The performance improvement ($\uparrow$) or drop  ($\downarrow$) with respect to the \textit{Ours} in Table \ref{table:acrossDataset} are given in parenthesis. Notice that higher values of AUC and lower values of Avg.Dist. mean an improvement. Results in black are better than its counterpart. Full refers to Eq.~\ref{eq:daFull}.}
    \centering
  %  \resizebox{0.7\linewidth}{!}
 %   {
    {\begin{tabular}{lcccc} \hline  \hline
    
        & Source & Target & AUC (\%) & {Avg.Dist.} \\ \hline 
        
       Ferreri et al. \small{(2021)}~\normalsize{~\cite{ferreri2021multi}} & GazeFollow & GOO & 66.1 (12.2$\downarrow$) & 0.327 (0.043$\downarrow$) \\ 
        \textbf{Ours} \small{(Full)} & GazeFollow & GOO & \textbf{84.0} (5.7$\uparrow$) & \textbf{0.238} (0.046$\uparrow$) \\ \hdashline
        
     Ferreri et al. \small{(2021)}~\normalsize{~\cite{ferreri2021multi}} & VideoAttentionTarget & GOO & 61.8 (7.3$\downarrow$) & 0.388 (0.114$\downarrow$) \\ 
     \textbf{Ours} \small{($L_{GRL}$)} & VideoAttentionTarget & GOO & {75.3} (6.2$\uparrow$) & {0.300} (0.026$\downarrow$) \\ 
     \textbf{Ours} \small{($L_{RGB \rightarrow Depth}$ + $L_{Depth \rightarrow RGB}$)} & VideoAttentionTarget & GOO & {69.9}  (0.8$\uparrow$) & {0.301} (0.027$\downarrow$) \\ 
        \textbf{Ours} \small{(Full)} & VideoAttentionTarget & GOO & \textbf{77.5} (8.4$\uparrow$) & \textbf{0.257} (0.017$\uparrow$) \\  \hdashline
        
       Ferreri et al. \small{(2021)}~\normalsize{~\cite{ferreri2021multi}} & GOO & GazeFollow & 62.5 (0.4$\downarrow$) & \textbf{0.412} (0.011$\downarrow$) \\
        \textbf{Ours} \small{(Full)}  & GOO & GazeFollow & \textbf{64.2} (1.3$\uparrow$) & 0.413 (0.012$\downarrow$) \\  \hdashline
        
       Ferreri et al. \small{(2021)}~\normalsize{~\cite{ferreri2021multi}} & GOO & VideoAttentionTarget & 69.2 (11.7$\uparrow$) & \textbf{0.325} (0.121$\uparrow$) \\
        \textbf{Ours} \small{(Full)}  & GOO & VideoAttentionTarget & \textbf{73.1} (15.6$\uparrow$) & 0.360 (0.086$\uparrow$) \\

        \hline \hline
    \end{tabular}
%    }
   }
    \label{table:domanadapt}
\end{table}

The proposed DA notably improves the performance: AUC (1.3-15.6\% more) and Avg.Dist. (0.017-0.086 less) compared to without applying DA. These improvements are higher than Ferreri et al.~\cite{ferreri2021multi} in all settings in terms of AUC. In terms of Avg.Dist., for GOO $\rightarrow$ VideoAttentionTarget, \cite{ferreri2021multi} performs better than the proposed DA while for the rest of the settings the proposed DA outperforms Ferreri et al.~\cite{ferreri2021multi}. Also, one can observe that the performance across domains does not always raise up by applying Ferreri et al.~\cite{ferreri2021multi} (e.g., GazeFollow $\rightarrow$ GOO and vice-versa).

\subsection{Qualitative Results}\label{sec:qual}
\begin{figure*}[th!]
    \centering
    \includegraphics[width=0.85\textwidth]{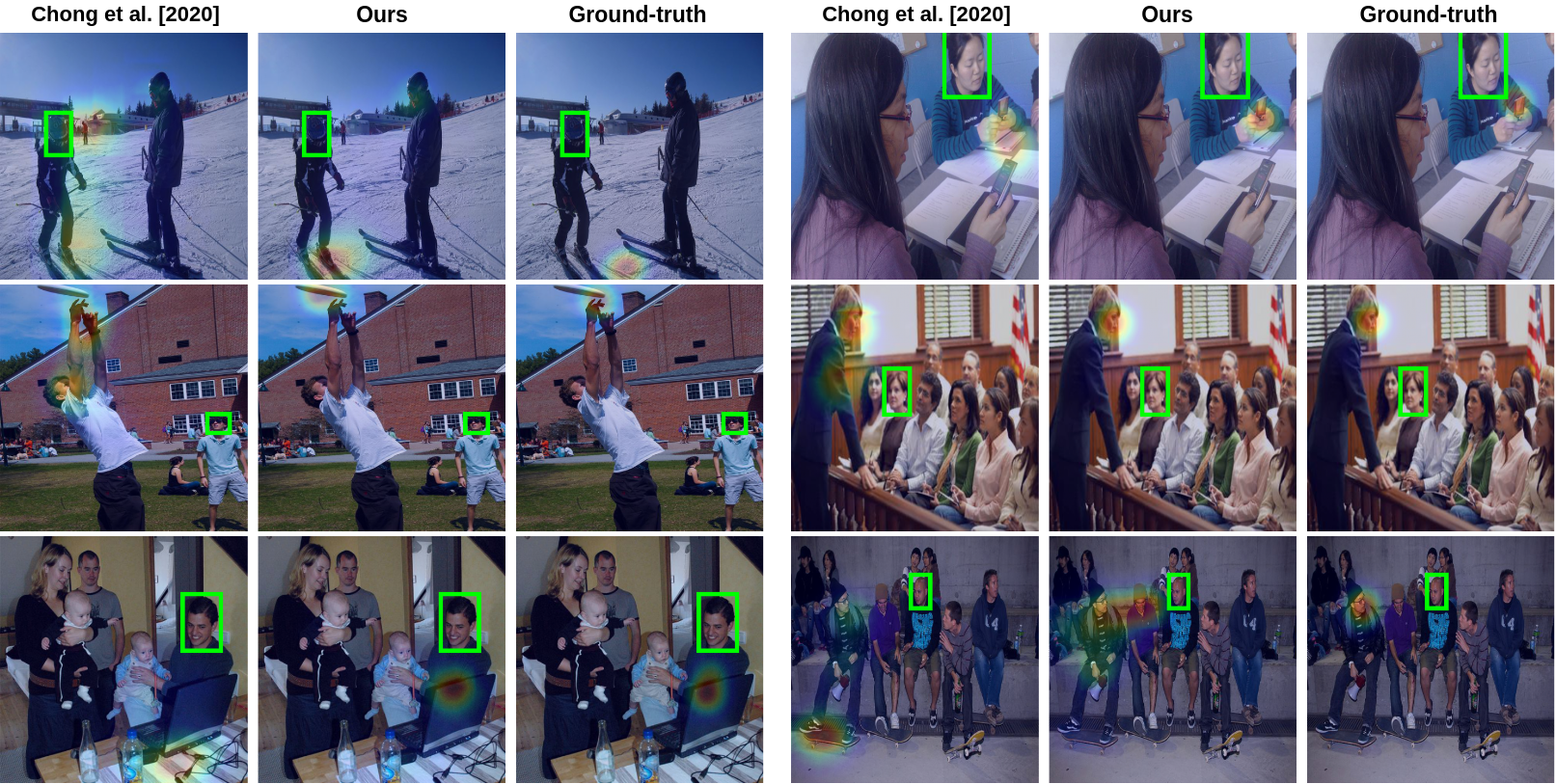}
    \caption{Qualitative results on the GazeFollow~\cite{Recasens2015} dataset. We evaluate the performance of our multimodal network, Chong et al. \cite{chong2020detecting} with respect to the ground-truth data. Green bounding boxes are taken from the corresponding dataset, referring to the cropped head image of the person whose gaze target to be detected.}
    \Description{Multiple images. Each contains multiple people in social interaction with each other as well as multiple objects and actions.}
    \label{fig:qualitative}
\end{figure*}

\begin{figure*}[th!]
    \centering
    \includegraphics[width=0.85\textwidth]{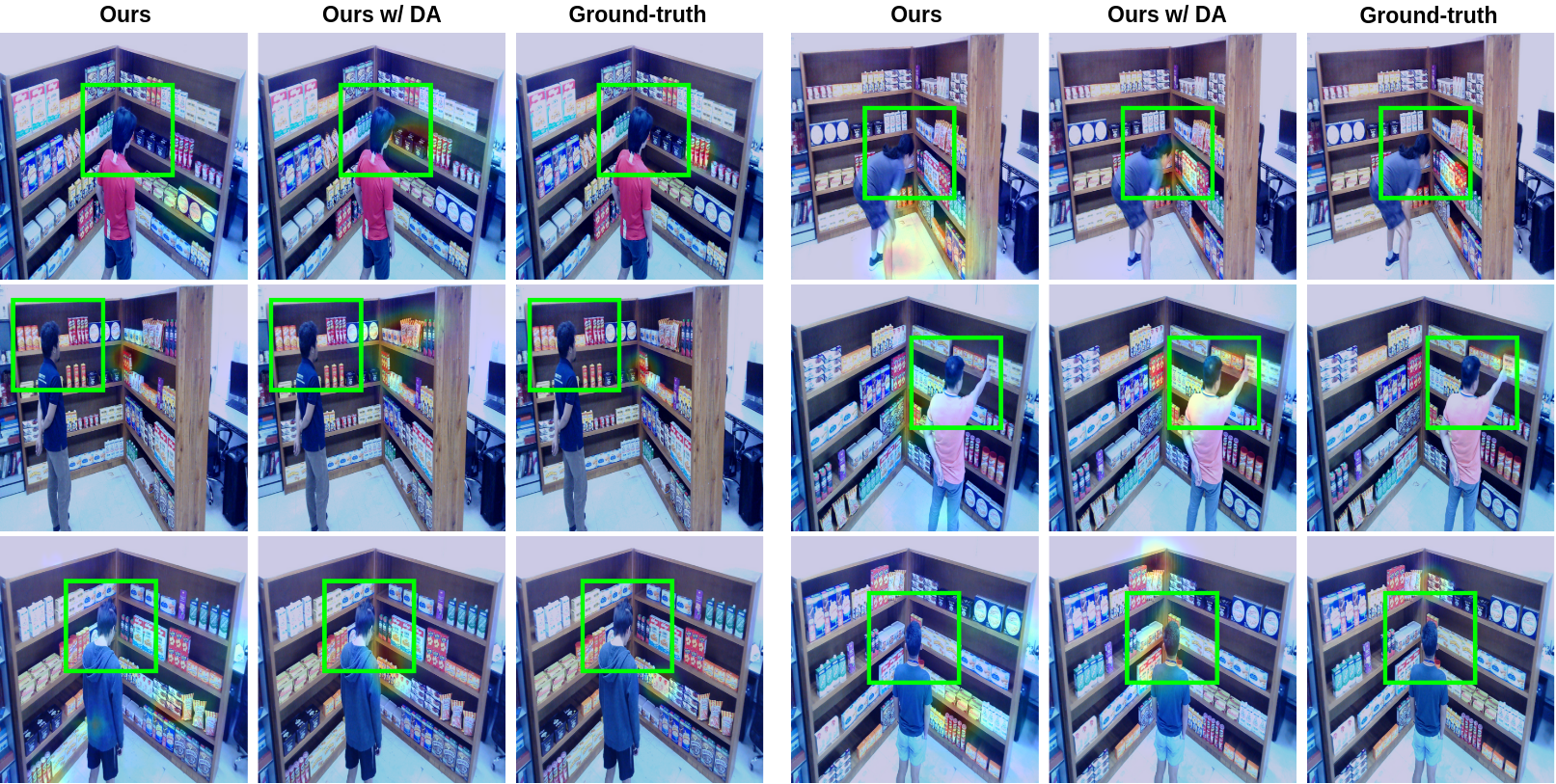}
    \caption{Qualitative results of our proposal with and without the domain adaptation component when the source data is GazeFollow~\cite{Recasens2015} and the target data is GOO~\cite{tomas2021goo}. Green bounding boxes are taken from the corresponding dataset, referring to the cropped head image of the person whose gaze target to be detected.}
    \Description{Multiple images of people shopping in a supermarket.}
    \vspace{-0.5cm}
    \label{fig:qualitative2}
\end{figure*}

Fig.~\ref{fig:qualitative} shows qualitative results on GazeFollow~\cite{Recasens2015} dataset, in which we present our multimodal network's (i.e., no DA) and Chong et al.~\cite{chong2020detecting}'s results together with the ground-truth. As seen, our neural network is able to capture gaze in challenging and dynamic scenes as well as producing more compact heatmaps (implying less Avg.Dist.) compared to Chong et al.~\cite{chong2020detecting}. In Fig.~\ref{fig:qualitative2}, we demonstrate results of our model with and without domain adaptation. The corresponding images are obtained when the source dataset is GazeFollow~\cite{Recasens2015} and the target dataset is GOO~\cite{tomas2021goo}. Given the ground-truth data, one can observe that our domain adaptation method notably improves the gaze target detection results compared to our multimodal network without domain adaptation.

\section{Discussion}
\label{sec:conclusion}
We have presented a novel multimodal deep architecture in order to identify where the person, in an image taken from the third perspective, is looking. Our spatial model is composed of three-pathways simultaneously processing \textit{i)} the head image belonging to the person whose gaze target is to be detected (i.e., person-of-interest), \textit{ii)} the scene image and \textit{iii)} the depth maps, both supplying the context information. It is distinguishable from the prior art as it does not rely on supervision of gaze angle, does not require explicit head orientation information or the location of the eyes of the person-of-interest.
Extensive quantitative and qualitative evaluations demonstrate that the proposed method performs favorably against the existing approaches. Additionally, our investigations regarding joint-learning of multiple modalities resulted in several variations of the proposed method. Some of these variations, notice that they have never presented in an earlier work, also outperforms several SOTA.

First time in this paper, we also studied the domain adaption (DA) for gaze target detection. To do so, we have injected new DA components to the described multimodal network. Our proposal enhanced the performance on target datasets as well as performing better than the DA SOTA. It is important to mention that the used datasets were all collected in unconstrained situations, including complex human-human social interactions and/or human-object interactions. The effective results of the proposed method on these benchmarks, and particularly its capability to handle domain-shift problem, potentially makes it stronger than the counterparts when it is integrated to real-life applications.
Inline with SOTA~\cite{chong2020detecting,fang2021dual}, the proposed method not only detects the gaze targets located in the scene but also able to declare if the gaze target is out-of-the-scene. In this paper, we have not evaluated our method in terms of out-of-frame precision~\cite{chong2020detecting,fang2021dual} (consequently, it is discarded from the contributions as well). This is due to lack of corresponding annotations in multiple datasets, which does not allow us to train and/or test our pipeline, particularly the DA component. 
One potential future work will be supplying out-of-frame annotations for all existing benchmarks (at least for their test splits). Additionally, the proposed method will be tested on unconstrained human-robot interaction scenarios targeting assistive robotics application in hospitals using the real-life dataset collected by the EU Horizon 2020 SPRING project (No. 871245). We will also investigate integrating transformers into our multimodal pipeline to better exploit the head attention over the scene and depth networks.

\begin{acks}
This work was supported by the EU Horizon 2020 SPRING project (No. 871245).
\end{acks}

\bibliographystyle{ACM-Reference-Format}
\bibliography{mybib}

\end{document}